\documentclass[lettersize,journal]{IEEEtran}
\usepackage{amsmath,amsfonts}
\usepackage{algorithmic}
\usepackage{algorithm}
\usepackage{array}
\usepackage[caption=false,font=normalsize,labelfont=sf,textfont=sf]{subfig}
\usepackage{textcomp}
\usepackage{stfloats}
\usepackage{url}
\usepackage{verbatim}
\usepackage{graphicx}
\usepackage{cite}
\usepackage{booktabs}
\usepackage{balance}
\usepackage{multirow}
\usepackage[switch,pagewise]{lineno}
\usepackage{color}
\usepackage{hyperref}
\hypersetup{colorlinks=true}

\begin{document}

\title{Lightweight Facial Attractiveness Prediction \\ Using Dual Label Distribution}

\author{Shu Liu, \IEEEmembership{Member, IEEE,} Enquan Huang, Ziyu Zhou, Yan Xu, Xiaoyan Kui, \\Tao Lei, \IEEEmembership{Senior Member, IEEE,} Hongying Meng, \IEEEmembership{Senior Member, IEEE}
\thanks{This work was supported by the National Natural Science Foundation of China under Grants U22A2034, 62177047 and 62271296, Hunan Provincial Natural Science Foundation of China under Grant 2023JJ30700, and Central South University Research Programme of Advanced Interdisciplinary Studies under Grant 2023QYJC020. We are grateful for resources from the High Performance Computing Center of Central South University.}

\thanks{Shu Liu, Enquan Huang, Ziyu Zhou, Yan Xu, and Xiaoyan Kui are with the School of Computer Science and Engineering, Central South University, Changsha 410083, Hunan, China (e-mail: \{sliu35, enquan, zzyotl, taylor\_xy0827, xykui\}@csu.edu.cn).}

\thanks{Tao Lei is with the School of Electronic Information and Artificial Intelligence, Shaanxi University of Science and Technology, Xi'an 710021, Shaanxi, China (e-mail: leitao@sust.edu.cn).}%
\thanks{Hongying Meng is with the Department of Electronic and Electrical Engineering, Brunel University London, Uxbridge UB8 3PH, UK (e-mail: hongying.meng@brunel.ac.uk).}%
}

\markboth{Journal of \LaTeX\ Class Files,~Vol.~XX, No.~X, 2024}%
{Shell \MakeLowercase{\textit{\textit{et al.}}}: A Sample Article Using IEEEtran.cls for IEEE Journals}

\IEEEpubid{0000--0000/00\$00.00~\copyright~2024 IEEE}

\maketitle

\begin{abstract}
   Facial attractiveness prediction (FAP) aims to assess {facial} attractiveness automatically based on human aesthetic perception. Previous methods using deep convolutional neural networks have {improved} the performance, but their large-scale models {have led} to a deficiency in flexibility. {In addition, most methods} fail to take full advantage of the dataset. In this paper, we present a novel end-to-end FAP approach {that integrates} dual label distribution and lightweight design. {The} manual ratings, attractiveness score, and standard deviation are aggregated explicitly to construct a {dual-label distribution to make the best use of the dataset}, including the attractiveness distribution and the rating distribution. Such distributions, as well as the attractiveness score, are optimized under a joint learning framework based on the label distribution learning (LDL) paradigm. The data processing is simplified to a minimum {for a lightweight design}, and MobileNetV2 is selected as our backbone. Extensive experiments are conducted on two benchmark datasets, where our approach achieves promising results and succeeds in {balancing} performance and efficiency. Ablation studies demonstrate that our delicately designed learning modules are indispensable and correlated. Additionally, the visualization indicates that our approach can perceive facial attractiveness and capture attractive facial regions to facilitate semantic predictions. The code is available at \url{https://github.com/enquan/2D\_FAP}.
\end{abstract}



\section{Introduction}

\IEEEPARstart{F}{acial} attractiveness plays a significant role in {daily life} \cite{bashour2006history,rhodes2011oxford}. It is a complex and multifactorial concept, devoting researchers from diverse disciplines to decrypting its mysteries \cite{ibanez2019subjectivity}. Two views from {the} social sciences have been debated. One is the {long-term} belief that beauty is in the eye of the beholder, indicating the culture-bound and personalized {properties} of facial attractiveness. The other is the common notion of general consensus in beauty {judgments} among observers, indicating its universal human preference. Although the controversy of subjectivity and universality in aesthetic perception {has not been} settled, these findings {form} the cognitive basis for facial attractiveness research in computer science \cite{liu2016advances}. Facial attractiveness prediction (FAP) aims to assess facial attractiveness automatically based on human perception, which facilitates the development of many real-life applications, such as face manipulation and retrieval \cite{chen2016data,ning2023face}, social media recommendation \cite{rothe2016some,abousaleh2021multimodal}, and cosmetic surgery \cite{bottino2012new}.


In the past two decades, FAP has gradually become a prosperous research topic in computer vision. The methods can be categorized into handcrafted feature based and deep learning based. In most early studies, low-level features like geometric \cite{aarabi2001automatic,chen2016combining} and texture descriptors \cite{kagian2007humanlike,zhang2016computer} {were} manually designed. Such a representation, however, may lack discriminative capability, resulting in poor performance. With the {emergence} of deep learning, convolutional neural networks (CNNs) \cite{gray2010predicting,lin2019regression,lin2019attribute} have been applied to FAP. Due to {their} powerful nonlinearity, CNN-based methods are able to learn hierarchical aesthetic {representations} thus boosting the performance, but their large-scale models {lack} flexibility. Adapting neural network architectures to strike a balance between performance and efficiency has been an active research field in recent years. Unfortunately, {the lightweight design of} FAP has {received} little attention. The only work utilizing lightweight backbone is to employ MobileNetV2 with co-attention learning mechanism~\cite{shi2019improving}.

\begin{figure*}[!t]
    \centering
    \includegraphics[width = .86\linewidth]{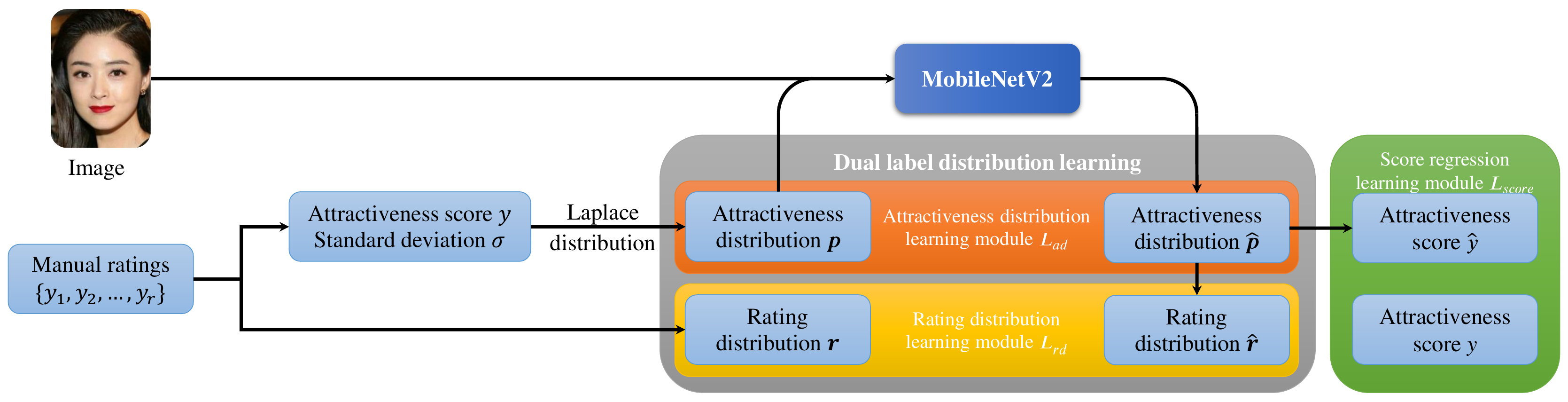}
      \caption{{Overview} of our framework. For a given facial image, its manual ratings, attractiveness score, and standard deviation are aggregated explicitly. With the human ratings $\{y_1,y_2,\dots,y_r\}$, its attractiveness distribution $\boldsymbol p$ is generated using the Laplace distribution, while its rating distribution $\boldsymbol r$ is derived directly. Then, the facial image and $\boldsymbol p$ are fed into MobileNetV2 to output the predicted attractiveness distribution $\hat {\boldsymbol{p}}$, which is subsequently utilized to compute the predicted attractiveness score $\hat y$ and obtain the predicted rating distribution $\hat{\boldsymbol r}$. Finally, $\hat {\boldsymbol{p}}$, $\hat{\boldsymbol r}$ and $\hat y$ are jointly optimized under the {dual-label} distribution and score regression learning modules.}
      \label{fig:framework}
\end{figure*}




\IEEEpubidadjcol

The FAP datasets usually include facial images with their corresponding manual ratings, ground-truth scores, and standard deviations. Therefore, among CNN-based methods, several labeling schemes have been adopted to meet different learning objectives and employ {datasets} to varying degrees. The single-label (average score) is the most commonly used, but only one type of label {is considered}, thus imposing strong restrictions {on} learning. Although the multi-label scheme cannot {adapt well} to FAP, its variant, label distribution learning (LDL), has been introduced to provide a novel view of attractiveness learning~\cite{fan2017label}. Human ratings are aggregated into {a} label distribution, while the ground-truth score and standard deviation {are not} utilized explicitly. The LDL paradigm, which aims to learn the latent distributions in the dataset, was formally proposed in~\cite{geng2016label}. Since then, it has been applied to various tasks, such as age estimation~\cite{gao2018age}, emotion distribution recognition~\cite{zhou2015emotion}, and facial landmark detection~\cite{su2019soft}.


In this paper, we present a novel end-to-end FAP approach consisting of {a dual-label} distribution and joint learning framework based on lightweight design. The lightweight design lies in many aspects, from data processing to backbone selection. We have simplified the data preprocessing and augmentation to minimum, and conducted extensive experiments to reach the final decision to employ MobileNetV2 as the backbone.

{The overview of our framework is shown in Fig. \ref{fig:framework}.} To make the best use of the dataset, the dual-label distribution, including the attractiveness and rating {distributions}, is proposed and constructed to {explicitly} utilize the manual ratings, ground-truth score, and standard deviation. Then, it is fed into MobileNetV2 for joint learning, which is designed to optimize three learning modules simultaneously based on the LDL paradigm. The attractiveness distribution learning module aims to optimize the network output, {namely,} the predicted attractiveness distribution. The rating distribution learning module is designed to refine the predicted rating distribution {while} further {supervising the} learning process. The score regression learning module concentrates on further refining the predicted attractiveness score with a novel loss. Finally, given a facial image, the trained model outputs its predicted attractiveness distribution and obtains its attractiveness score to accomplish end-to-end FAP.




The contributions of this paper are summarized as follows.
\begin{itemize}
    \item {We present a novel end-to-end FAP approach that is the first to leverage the lightweight design and LDL paradigm to predict facial attractiveness.}
    \item {A dual-label} distribution is proposed to take full advantage of the dataset, including the manual ratings, ground-truth score, and standard deviation.
    \item {A} joint learning framework is further proposed to optimize the {dual-label} distribution and concurrently refine the predictions using a novel loss.
    \item Extensive experiments are conducted on two benchmarks, where our approach achieves appealing results with greatly decreased parameters and {computations}.

\end{itemize}

\section{Related Work}

\subsection{Deep Learning Based Facial Attractiveness Prediction}

The handcrafted feature based methods advanced the {FAP} field and {achieved} some early success. However, such methods suffer from multiple limitations. First, they are dependent on low-level features {that} lack representational capability, leading to inferior performance. Second, the model performance relies heavily on {feature selection, which the process} can be complicated and highly empirical. Third, handcrafted features are strongly constrained because most are based on existing aesthetic criteria or findings from psychological research.

With the {emergence} of deep learning, {numerous} CNN-related works have been proposed, many of which have obtained remarkable results on some challenging visual classification or recognition tasks \cite{krizhevsky2012imagenet,Parkhi15}. Meanwhile, the effectiveness of {CNNs and their variants has} been extensively explored in facial attractiveness prediction. Gray \textit{et al.} were the first to construct a CNN-like hierarchical feedforward model to extract attractiveness features for FAP task \cite{gray2010predicting}. A six-layer CNN {was designed} to learn the features {at} multiple levels and directly output the attractiveness score \cite{xie2015scut}. Later, {the} VGG network \cite{simonyan2014very} was proposed to extract discriminative deep facial features, which were {subsequently} utilized for attractiveness prediction \cite{xu2018transferring}. To a certain extent, deeper architectures enable the extraction of more discriminative and representative features.

Simultaneously, researchers have been seeking other paths for enhanced performance and more general solutions. A {psychologically inspired} CNN (PI-CNN) \cite{xu2017facial} {was proposed} for FAP {that was} fine-tuned with different aesthetic features. Inspired by the effectiveness of facial attributes on facial attractiveness, an attribute-aware CNN (AaNet) \cite{lin2019attribute} {was proposed} that can integrate attractiveness-related attributes into feature representation to {adaptively modulate network filters}. Most recently, {FAP was redefined} as a {ranking-guided} regression task, {where a ranking-guided} CNN (R$^3$CNN) {was constructed to} accomplish ranking and regression simultaneously \cite{lin2019regression}. A two-branch architecture named REX-INCEP {was presented in} \cite{bougourzi2022deep}. {It employed} multiple dynamic loss functions {and} established an ensemble regressor (CNN-ER) for FAP, which {comprises} 6 models involving the proposed REX-INCEP. In addition, {FAP} can be {combined} with other visual tasks. A hierarchical multitask network that can {concurrently determine} gender, race, and facial attractiveness of a given portrait image {has been designed} \cite{xu2019hierarchical}. Recently, a multitask FAP model {was also introduced} to automatically {predict} facial attractiveness and gender \cite{xu2021mt}.

It is worth noting that FAP research aims to mimic human attractiveness perception, including universal preference shared by a diverse group of observers, and personalized preference of particular individuals. This paper {focuses} on the universal facial attractiveness prediction, which is also the scope of most existing studies.

\subsection{Lightweight Architecture}

{Model efficiency} is often a vital indicator in deep learning tasks {and} is measured by the number of trainable parameters, floating point operations per second (FLOPs), and multiply-adds (MAdds) \cite{sandler2018mobilenetv2}. In recent years, extensive studies have {attempted} to adapt neural network {architectures to balance} between model efficiency and performance, i.e., reducing the amount of model parameters or MAdds while maintaining relatively high performance in multiple tasks, such as SqueezeNet \cite{iandola2016squeezenet}, EfficientNet \cite{tan2019efficientnet}, and MobileNet. The MobileNet variants largely depend on separable convolutions to decrease the model size, which decompose standard convolutions into a 1$\times$1 pointwise convolution and a depthwise convolution applied to each channel separately. MobileNetV1 \cite{howard2017mobilenets} is based on a streamlined architecture {that uses depthwise} separable convolutions. MobileNetV2 \cite{sandler2018mobilenetv2} utilizes the proposed linear bottleneck with an inverted residual structure. Later, MobileNetV3 \cite{howard2019searching} was developed through hardware-aware network architecture search, which adopts squeeze and excitation and nonlinearities like swish. Recently, Zhou \textit{et al.} \cite{zhou2020rethinking} {analyzed the disadvantage} of the inverted residual block in MobileNetV2, {and} presented a bottleneck {known as the} sandglass block. {It was then} employed to construct MobileNeXt.


The aforementioned lightweight architectures have been broadly employed in multiple tasks, {such as} popular classification and object detection. A concise self-training method for ImageNet classification {was presented}, iteratively training smaller and larger EfficientNet models {as student and teacher, respectively} \cite{xie2020self}. A novel family of object detectors named EfficientDet {was developed} based on EfficientNet backbones and several optimizations for object detection, including a weighted bidirectional feature pyramid network and a compound scaling method \cite{tan2020efficientdet}. In addition to employing lightweight backbones directly, many researchers have been dedicated to {developing} models for various tasks based on the lightweight design. The separable convolutions proposed in MobileNets {were transferred} to construct an efficient graph convolutional network for skeleton-based action recognition \cite{song2022constructing}. Most recently, the first layer and the first convolutional linear bottleneck of MobileNetV2 {were borrowed} as feature extractors in a proposed lightweight single-image segmentation network \cite{sun2022gaussian}.


Although lightweight design has been widely adopted in many tasks, {it has been} largely ignored in FAP. {Only} few related studies have been carried out. The prediction {of facial attractiveness was enhanced} by utilizing {pixelwise} labeling masks for accurate facial composition {and a} co-attention learning mechanism with MobileNetV2 \cite{shi2019improving}.


\subsection{Label Distribution Learning}

Previous learning paradigms, such as single-label learning (SLL) and mutli-label learning (MLL), address the fundamental question of which label describes the instance. However, neither SLL nor MLL can directly handle {further questions} with more ambiguity. They {are not suitable for} some real applications in which the overall distribution of the labels matters. Besides, real-world data with natural measures of {label importance exist}. Motivated by the above facts, label distribution learning was formally proposed by Geng in 2016 \cite{geng2016label}, which is a more general learning framework than SLL and MLL. It concentrates on the ambiguity on the label side to learn the latent distribution of the labels. {Generally,} the label distribution involves a certain number of labels, each {describing} the importance to the instance.

{LDL} has been adopted in a wide range of tasks. The emotion distribution learning method {was proposed} to output the intensity of all basic emotions on a given image \cite{zhou2015emotion}. {This method addresses} the issue of treating the facial expression {in} an image as {only} a single emotion. One unified framework {with} a lightweight architecture {was designed} to jointly learn age distribution and regress age using the expectation of age distribution \cite{gao2018age}. {This approach alleviates the high computational cost} of large-scale models and the inconsistency between the training and evaluation phases. Motivated by inaccurately annotated landmarks, the soft facial landmark detection algorithm {was developed in} \cite{su2019soft}. It associates each landmark with a bivariate label distribution (BLD), learns the mappings from an image patch to the BLD for each landmark, and finally obtains the facial shape based on the predicted BLDs.

In addition to the aforementioned works, {LDL} has also been employed in the FAP field. Ren and Geng \cite{ren2017sense} proposed a beauty distribution transformation to convert $k$-wise ratings to label distribution, and a structural LDL method based on structural support vector machine to reveal the human sense of facial attractiveness. Our previous work \cite{fan2017label} utilized the inherent score distribution of each image given by human raters as the learning objective and integrated low-level geometric features with high-level CNN features to accomplish automatic attractiveness computation. A deep adaptive LDL framework was developed by Chen and Deng, utilizing discrete label distribution of possible ratings to supervise the FAP learning process \cite{chen2019facial}. Later, the deep label distribution learning-v2 (DLDL-v2) approach, which {originated} from age estimation \cite{gao2018age}, {was further designed} to estimate facial attractiveness based on the expectation of label distribution through the lightweight ThinAttNet and TinyAttNet \cite{gao2020learning}.

\section{Dual Label Distribution}\label{sec:dld}

In this section, we present the construction of the {dual-label} distribution. Some preliminaries are {firstly} introduced to lay the foundation. Then, the {dual-label} distribution is proposed in detail, including the rating distribution and the attractiveness distribution, which construct an actual rating distribution and a pseudo rating distribution, respectively. Such distributions are complementary by utilizing the variety of information in the dataset, thus facilitating the network learning.

\subsection{Preliminaries}

\subsubsection{The Facial Attractiveness Prediction Problem}

Assume that a training set with $N$ samples is denoted as $\{(x^{(i)},y^{(i)})\}^N_{i=1}$, where $x^{(i)}$ and $y^{(i)}$ denote the $i$-th image and its ground-truth score (average score), {respectively}. We might omit the superscript $(i)$ for simplicity. The goal of {FAP} is to learn a mapping from facial images to attractiveness scores such that the error between the predicted score $\hat y$ and ground-truth score $y$ {is} as small as possible on an input image $x$.


\subsubsection{Laplace Distribution}

Both the Laplace distribution and Gaussian distribution are widely used in statistics and data analysis. They share {similar symmetry, continuity, and peak location properties}. When the variance {in} the Gaussian distribution approaches infinity, it becomes the Laplace distribution. In this sense, the Laplace distribution can be seen as a more general distribution that includes the Gaussian distribution as a special case. Besides, {calculating} the probability distribution function for the Laplace distribution is simpler than {that for} the Gaussian distribution. Therefore, we choose the Laplace distribution in the construction of attractiveness distribution. Defined by the location parameter $\mu$ and the scale parameter $b$ (their settings for each image {are} introduced in Section \ref{sec:attractiveness}), the probability density function of the Laplace distribution is


\begin{equation}
    f(x|\mu,b)=\frac{1}{2b}\exp(-\frac{|x-\mu|}b)
\end{equation}
while the cumulative distribution function is
\begin{equation}
    F(x|\mu,b)=\frac 1 2 [1+\text{sgn}(x-\mu)(1-\exp(-\frac {|x-\mu|}b))]
\end{equation}
where the mean and standard deviation of the distribution are $\mu$ and $\sqrt 2 b$, respectively.

During {the} training phase, the attractiveness distribution is fed into the model, and the rating distribution is employed for supervision. Their generation is introduced in the following.



\subsection{Rating Distribution}\label{sec:rating}

In our previous work\cite{fan2017label}, a {LDL-based FAP} method was proposed, which utilized the rating records directly to derive the rating distribution. The ground-truth score and standard deviation, however, were implicitly included, thus ignoring their importance. We follow the same way to construct the rating distribution, represented by the vector $\boldsymbol r$.

Let $r_m$ be the number of raters who rated the image with the attractiveness score $m$. Since the attractiveness score is an integer ranging from 1 to 5, $m=\{1,2,3,4,5\}$. Then $L_1$ normalization is applied to $\boldsymbol r$ such that $\sum_{m=1}^{5}r_m=1$. {Thus,} $\boldsymbol r$ represents the actual rating distribution of the image.

\subsection{Attractiveness Distribution}\label{sec:attractiveness}

In order to utilize the ground-truth score $y$ and the standard deviation $\sigma$ of the image explicitly, the attractiveness distribution are constructed, which takes advantage of LDL and is represented by the vector $\boldsymbol p$.

The formation of $\boldsymbol p$ is introduced as follows. Each element of $\boldsymbol p$ represents the probability of the attractiveness score on a certain interval. These probabilities are then combined to establish the attractiveness distribution. First, we define the interval endpoints $s_k$
\begin{equation}
    s_k=y_{\min}+k\cdot \Delta l
    \label{eq:sk}
\end{equation}
where $y_{\min}$ and $\Delta l$ are the minimum attractiveness score and interval length, respectively.

Then, the $j$-th interval $I_j$ is formed as
\begin{equation}
    I_j=[s_j,s_{j+1}]
    \label{eq:ij}
\end{equation}
Its corresponding probability $p_j$, {namely,} the $j$-th element of $\boldsymbol p$, is calculated using the cumulative distribution function of the Laplace distribution $F(x|\mu,b)$.
\begin{equation}
    p_j=F(s_{j+1}|\mu,b)-F(s_j|\mu,b)
    \label{eq:pj}
\end{equation}
where the location parameter and scale parameter are set to $\mu=y$ and $b=\frac{\sigma}{\sqrt{2}}$ for each image, {respectively. It} is consistent with the mathematical definition, thereby the construction is logical and expected to be viable.

In this work, the interval length $\Delta l$ is 0.1. To make the best choice of $\Delta l$, we have conducted a series of comparative experiments with $\Delta l=\{0.01,0.05,0.1,0.2,0.5\}$, and found that either {a} larger or smaller $\Delta l$ would {negatively impact the} performance. {Specifically}, with {a} larger $\Delta l$, the model outputs {a} sparser representation of the attractiveness distribution, which directly decreases its precision and further affects the derived rating distribution. {In contrast}, with {a} smaller $\Delta l$, the model has to output {a} distribution with higher dimensions, which is definitely a challenge for lightweight architecture due to its limited representational power.

Since the attractiveness score ranges from 1 to 5 in our adopted datasets, $s_k$ should share the identical range. Let $y_{\max}$ and $y_{\min}$ be the maximum and minimum attractiveness {scores, respectively; then,} $y_{\min}=1$ and $y_{\max}=5$. {Note} that $s_k$, $I_j$, and $p_j$ are 0-indexed. We have $k_{\max}=(y_{\max}-y_{\min})/\Delta l=40$. Hence, in Eqs. (\ref{eq:sk})-(\ref{eq:pj}), $k\in[0,40],j\in[0,39]$.

Finally, we perform {an elementwise} sigmoid operation and $L_1$ normalization to $\boldsymbol p$. The sigmoid operation offers nonlinear variation to $\boldsymbol p$, enhancing its representational power. The $L_1$ normalization is performed such that $\sum_{j=0}^{39}p_j=1$, hence satisfying the general property of a probability distribution.

\section{Learning}

In this section, the employed network architecture and the proposed joint learning framework are introduced in detail.

\subsection{Network Architecture}

\begin{figure}[t]
    \centering
    \includegraphics[width = \linewidth]{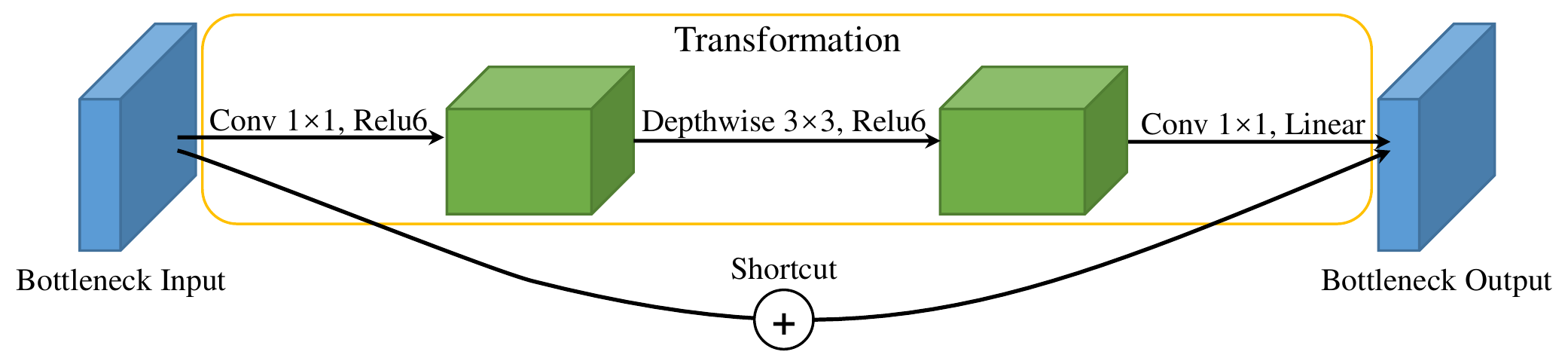}
    \caption{The building block of MobileNetV2, which consists of linear bottlenecks and an inverted residual structure.}
    \label{fig:bottleneck}
\end{figure}

Considering the balance between performance and efficiency, MobileNetV2 \cite{sandler2018mobilenetv2} is selected as our backbone after extensive experiments. As Fig. \ref{fig:bottleneck} {shows}, the building block of MobileNetV2 includes a 1$\times$1 expansion convolution {and} depthwise convolutions followed by a 1$\times$1 projection layer. The narrow input and output (bottleneck) are connected with a residual connection. This structure greatly reduces {the} model size and {number of computations} while maintaining relatively high performance on multiple tasks. Besides, sigmoid operation and $L_1$ normalization are also conducted on the output {to} reduce the inconsistency between the input and output.

\subsection{Joint Learning Framework}

The joint learning framework contains {an} attractiveness distribution, {a} rating distribution and score regression learning modules. As Fig.~\ref{fig:framework} {shows}, the input image {and} its attractiveness distribution are fed into MobileNetV2 to jointly optimize {the dual-label} distribution and attractiveness score in an end-to-end manner.


\subsubsection{Attractiveness Distribution Learning}

The Kullback-Leibler divergence is commonly used {in measuring the difference between} two probability distributions ~\cite{fan2017label,gao2018age}. However, we adopt {the} Euclidean distance to measure the similarity between $\boldsymbol p$ and its prediction $\hat{\boldsymbol{p}}$, whose calculation is much simpler with even better performance. We propose the attractiveness distribution learning module by defining its loss $L_{ad}$. The parameter $n$ in the following equations denotes the number of samples in a minibatch.


\begin{equation}
    L_{ad}=\frac 1 n \sum_{i=1}^n\Vert\hat{\boldsymbol p}^{(i)}-\boldsymbol p^{(i)}\Vert_2
\end{equation}

\subsubsection{Rating Distribution Learning}

With a single learning module, the proposed approach is unable to perform well due to {a lack of model supervision}. Therefore, we introduce a rating distribution learning module to reinforce the learning process. The generation of the predicted rating distribution vector $\hat {\boldsymbol r}$ is described as follows.

Similar to the definition of $\boldsymbol r$, $\hat r_m$ represents the predicted probability of rating $m$, which can be derived from $\hat{\boldsymbol p}$ using clustering and the rule of rounding. For example, the predicted probability of rating 2 can be computed by $\hat {\boldsymbol p}$ on the interval $[1.5,2.5)$. In this sense, we can establish a mapping among the score intervals, ratings, and subscripts in $\hat {\boldsymbol p}$, {as} shown in Table \ref{tab:mapping}. Thus, $\hat r_m$ is defined as
\begin{equation}
    \begin{aligned}
        \hat r_m=\sum \hat p_j, j\in\begin{cases}
            [0,4], & m=1\\
            [10m-15,10m-6], & m=2,3,4\\
            [35,39], & m=5
        \end{cases}
    \end{aligned}
\end{equation}



With $\hat{\boldsymbol r}$ and $\boldsymbol r$, we can further supervise the training via the rating distribution loss $L_{rd}$. Once again, {the} Euclidean distance is used to measure the similarity.

\begin{equation}
    L_{rd}=\frac 1 n \sum_{i=1}^n\Vert\hat{\boldsymbol r}^{(i)}-\boldsymbol r^{(i)}\Vert_2
\end{equation}

\begin{table}[t]
    \centering
    \caption{The mapping among the score intervals, ratings, and subscripts in $\hat {\boldsymbol p}$.}
    \label{tab:mapping}
    \begin{tabular}{rrr}
    \hline
    Score interval & Rating & Subscript in $\hat {\boldsymbol{p}}$\\
    \hline
    $[1.0,1.5)$               & 1               & $[0,4]$                           \\
    $[1.5,2.5)$               & 2             & $[5,14]$                           \\
    $[2.5,3.5)$               & 3             & $[15,24]$                           \\
    $[3.5,4.5)$               & 4             & $[25,34]$                           \\
    $[4.5,5.0]$               & 5               & $[35,39]$                           \\\hline
    \end{tabular}
    \end{table}

\subsubsection{Score Regression Learning}

The above two modules can learn {dual-label} distribution but fail to {account for} the attractiveness score. Besides, inconsistency still exists between the training and evaluation stages. Therefore, it is natural to incorporate a score regression learning module to further advance the prediction.

The attractiveness score is regressed by
\begin{equation}
\begin{aligned}
    \hat y^{(i)}&=\sum_{j=0}^{39}w_j\hat p_j^{(i)}
\end{aligned}
\end{equation}
where $w_j$ is the midpoint of the score interval $[s_j,s_{j+1}]$ in $\hat {\boldsymbol p}$, {i.e.,} $w_j=\frac12(s_j+s_{j+1})$. The score regression is similar to the calculation of expectation, when $w_j$ and $p_j$ are seen as the weight and probability, respectively.

\begin{figure}[t]
    \centering
    \includegraphics[width = 0.7\linewidth]{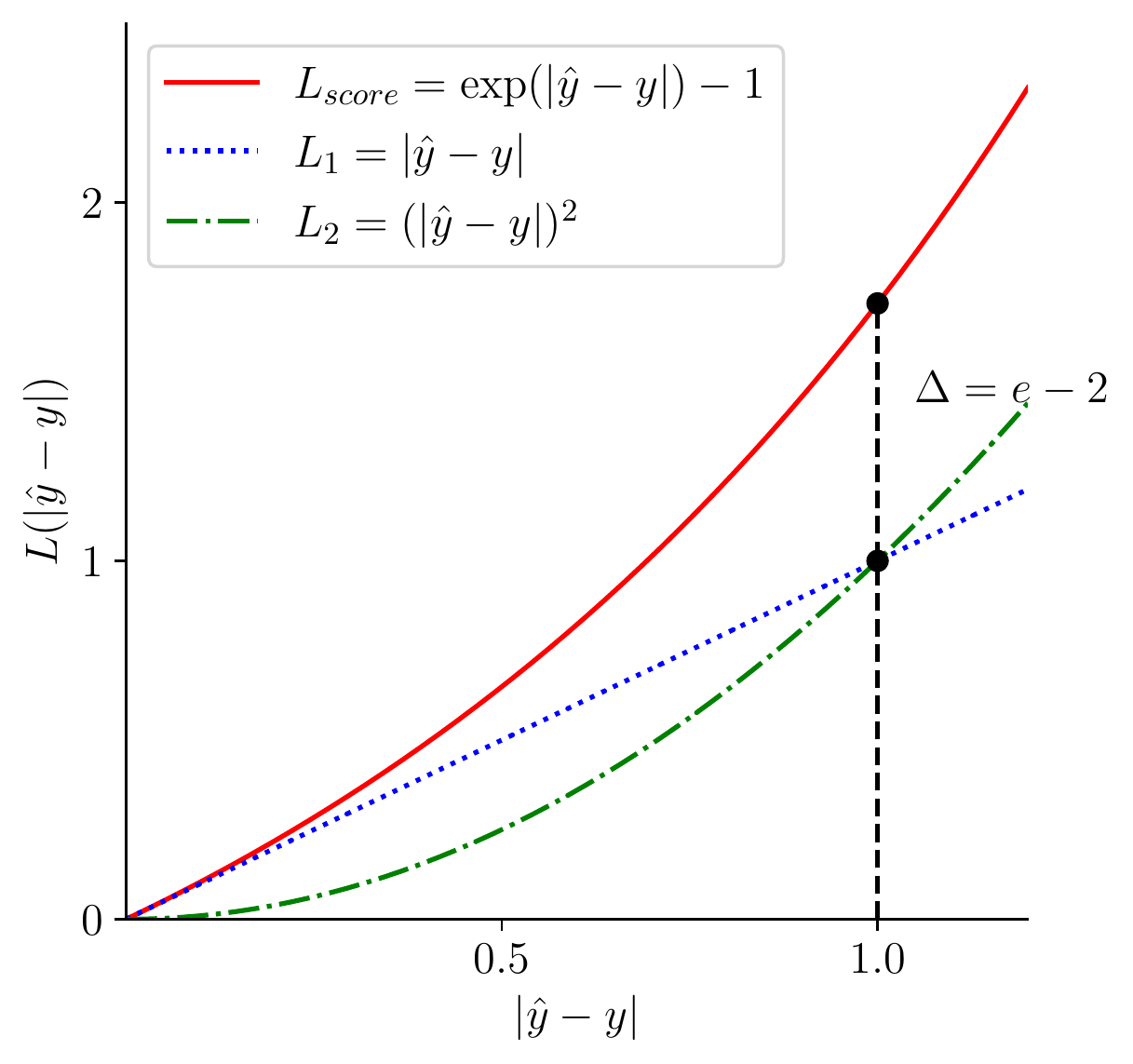}
    \caption{Comparison of the proposed $L_{score}$ and $L_1$, $L_2$ loss. When the absolute error between the ground-truth and predicted score reaches 1, $L_{score}$ is $e-2$ larger than $L_1$ or $L_2$ loss, thus refining the score prediction more vigorously.}
    \label{fig:loss}
\end{figure}

The score regression loss $L_{score}$ is then defined in Eq. (\ref{eq:score}). It is a combination of $\text{e}^v-1$ and $L_1$ loss, which is inspired by an important equivalent infinitesimal $\text{e}^v-1\sim v(v\to 0)$. Thus, $v$ is replaced by $\text{e}^v-1$. As shown in Fig.~\ref{fig:loss}, due to the property of exponential explosion, $L_{score}$ is much more sensitive to the difference between $\hat y$ and $y$ than $L_1$ or $L_2$ loss. The experiments also demonstrate such superiority.

\begin{equation}
    L_{score}=\sum_{i=1}^n[\exp(|\hat y^{(i)}-y^{(i)}|)-1]
    \label{eq:score}
\end{equation}

\subsubsection{Joint Loss}

The learning goal of our framework is to find the parameters {by jointly learning the} attractiveness distribution, rating distribution and score regression, so as to minimize the joint loss $L$.


\begin{equation}
    L=\lambda_1 L_{ad}+\lambda_2 L_{rd}+\lambda_3 L_{score}
\end{equation}
where $\lambda_1$, $\lambda_2$ and $\lambda_3$ are weights balancing the significance among {the} three types of losses.

Ablation studies with various combinations of $\lambda_i=\{1,2,5,10\}(i=1,2,3)$ have been performed. Intuitively, we expect the model with higher-weighted $\lambda_3$ {to perform} better because our task is to predict the attractiveness score, and the model should focus more on the score regression module. Surprisingly and coincidentally, the model with weights all set to 1 has the best overall performance; {thus}, $\lambda_1=\lambda_2=\lambda_3=1$ in our experiments.

\section{Experiments}

In this section, we present the experiments to validate the effectiveness of the proposed approach on two benchmark datasets. The implementation details, and comparisons with state-of-the-arts are thoroughly analyzed. {Afterward}, extensive ablation studies and visualization are carried out to further demonstrate the superiority of our approach.




\subsection{Implementation Details}

All the experiments are conducted on the popular deep learning framework PyTorch \cite{paszke2019pytorch} with an NVIDIA Tesla V100 GPU. To ensure the effectiveness, {five-fold} cross validation is performed, and the average results are reported.

\paragraph{Datasets} The SCUT-FBP5500 \cite{liang2018scut} and SCUT-FBP \cite{xie2015scut} datasets are used in our experiments. During the development of these datasets, volunteers were asked to rate {images} with integers ranging from 1 to 5, where the score 5 indicates the most attractive. Each image was then labeled with its average score. The full rating records are also provided, allowing us to construct different label distributions.

\paragraph{Data Preprocessing} Since the image size of SCUT-FBP dataset varies, we adopt {multitask} cascaded CNN (MTCNN) \cite{zhang2016joint} for face and facial landmark detection. Then, the faces are aligned to upright based on the detected landmarks and resized to 350$\times$350. No preprocessing {is performed} on SCUT-FBP5500 dataset. Finally, the images of SCUT-FBP5500 are resized to 256$\times$256 and center-cropped to 224$\times$224, while the aligned images of SCUT-FBP are resized to 224$\times$224 directly. Before feeding into the network, all resized images are normalized using the mean and standard deviation of the ImageNet dataset \cite{deng2009imagenet} for each color channel.

\paragraph{Data Augmentation} We only perform random horizontal {flipping with a probability of} $p=0.5$ in the training phase. No data augmentation {is} performed during {the} inference stage.


\paragraph{Training Details} We utilize the ImageNet-pretrained model to initialize the network. Then, the number of output channels of the fully-connected layer in the classifier module is modified to 40. The network is optimized by AdamW \cite{loshchilov2018decoupled}, with $\beta=(0.9,0.999)$ and $\epsilon=10^{-8}$. The initial learning rate is 0.001, and it is decreased by a factor of 10 every 30 epochs. Each model is trained {for} 90 epochs with a batch size of 256.

\paragraph{Inference Details} During {the} inference phase, the preprocessed images are fed into the network {to evaluate} our approach.

\paragraph{Evaluation Metrics} FAP can be formulated as a regression problem. Hence, the Pearson correlation coefficient (PC), mean absolute error (MAE), and root mean squared error (RMSE) are employed to evaluate the performance of our method, whose formulas are presented in Eq. (\ref{eq:metric}). A higher PC, {and} lower MAE and RMSE {suggest} better performance. Besides, the model efficiency is measured by the number of parameters and MAdds \cite{sandler2018mobilenetv2}.

\begin{equation}
    \begin{aligned}
        &\text{PC}=\frac{\sum_{i=1}^N(y^{(i)}-\bar y)(\hat y^{(i)}-\bar{\hat y})}{\sqrt{\sum_{i=1}^N(y^{(i)}-\bar y)^2}\sqrt{\sum_{i=1}^N(\hat y^{(i)}-\bar{\hat y})^2}}\\
        &\text{MAE}=\frac 1 N \sum_{i=1}^N|\hat y^{(i)}-y^{(i)}|\\
        &\text{RMSE}=\sqrt{\frac1N \sum_{i=1}^N(\hat y^{(i)}-y^{(i)})^2}
    \end{aligned}
    \label{eq:metric}
\end{equation}
where $N$ denotes the number of images in {the} test set, $\bar y=\frac1N\sum_{i=1}^Ny^{(i)}$, and $\bar{\hat y}=\frac1N\sum_{i=1}^N\hat y^{(i)}$.

\subsection{Comparison with the State of the Art}

We compare our approach against several recent and representative works on both datasets. The comparison with state of the art is given in Table \ref{tab:result}, which proves our advantages in terms of performance and efficiency. Note that all the comparison methods use the code and experimental settings {that are} publicly available in the original papers.

\begin{table*}[t]
    \centering
    \caption{Comparison with the state of the art on SCUT-FBP5500 and SCUT-FBP datasets. The best results are presented in bold.}
    \label{tab:result}
    \begin{tabular}{llrrrrr}
    \hline
    Method on SCUT-FBP5500 dataset               & Backbone    & \#Params(M) & MAdds(G) & PC$\uparrow$              & MAE$\downarrow$             & RMSE$\downarrow$            \\ \hline

    AaNet~\cite{lin2019attribute}                 & ResNet-18   & 11.69       & 1.82     & 0.9055          & 0.2236          & 0.2954          \\


    Co-attention learning~\cite{shi2019improving} & MobileNetV2$\times 2$ & 7.00           & 0.62     & 0.9260           & 0.2020           & 0.2660           \\

    MT-ResNet\cite{xu2021mt} & ResNet-50 & 25.56 & 4.11 & 0.8905 & 0.2459 & 0.3208\\

    R$^3$CNN~\cite{lin2019regression}                 & ResNeXt-50  & 25.03       & 4.26     & 0.9142          & 0.2120          & 0.2800          \\

    CNN-ER~\cite{bougourzi2022deep} & Ensemble of 6 models & 255.00 & - & 0.9250 & 0.2009 & 0.2650\\

    Ours                  & MobileNetV2 & \textbf{2.28}        & \textbf{0.31}     & \textbf{0.9276} & \textbf{0.1964} & \textbf{0.2585} \\ \hline \hline

    Method on SCUT-FBP dataset                 & Backbone    & \#Params(M) & MAdds(G) & PC$\uparrow$     & MAE$\downarrow$    & RMSE$\downarrow$   \\ \hline


    LDL~\cite{fan2017label} & ResNet-50   & 25.56       & 4.11     & 0.9301 & 0.2127 & 0.2781\\

    P-AaNet~\cite{lin2019attribute}                 & ResNet-18   & 11.69       & 1.82     & 0.9103 & 0.2224 & 0.2816 \\


    DLDL-v2 \cite{gao2020learning}& ThinAttNet & 3.69 & 3.86 & 0.9300 & \textbf{0.2120} & \textbf{0.2730}\\

    R$^3$CNN~\cite{lin2019regression}                   & ResNeXt-50  & 25.03       & 4.26     & \textbf{0.9500} & 0.2314 & 0.2885 \\

    Ours                    & MobileNetV2 & \textbf{2.28}        & \textbf{0.31}     & 0.9309 & 0.2212 & 0.2822 \\
    \hline
    \end{tabular}
    \end{table*}

\subsubsection{High performance}
{On} the SCUT-FBP5500 dataset, our approach achieves state-of-the-art performance on three evaluation metrics, surpassing previous methods utilizing non-lightweight backbones (e.g. ResNets, ResNeXts) \cite{lin2019attribute,xu2021mt,lin2019regression} by a large margin. The recently proposed CNN-ER \cite{bougourzi2022deep} adopted a large-scale ensemble model with dynamic loss functions to facilitate the prediction, yet our method performs slightly better. Additionally, when compared with the method using the same backbone \cite{shi2019improving}, ours still performs {slightly} better. {On} the SCUT-FBP dataset, our approach {yields} comparable results. Previous studies tend to adopt complicated data augmentation \cite{fan2017label,lin2019attribute} to seek better performance, while ours abandons such techniques to make  {them} simple and lightweight. As for DLDL-v2 \cite{gao2020learning}, we infer that its superiority largely resulted from pretraining. ThinAttNet was pretrained on the MS-Celeb-1M dataset, a face recognition dataset {that} is more relevant to our task than the object classification dataset, {e.g.,} the ImageNet dataset. Compared with the most recent R$^3$CNN \cite{lin2019regression}, although our approach has {a} slightly lower PC, its relatively higher MAE and RMSE {indicate} poorer predictions, which is unacceptable in FAP.


\subsubsection{High efficiency}
Our approach has the {fewest} parameters and MAdds among the compared methods. It is also an extension of our previous work \cite{fan2017label}, {as it significantly improves} model efficiency while achieving similar performance on SCUT-FBP. Compared with those using ResNet-18 or ResNet-50/ResNeXt-50, our method has {an} 80\% or 90\% reduction in the number of parameters and MAdds, respectively. A majority of previous methods employ deep models to {achieve} better performance. We, however, focus on the lightweight design to enable the model {to be} suitable for resource-constrained circumstances.

Overall, our approach succeeds in striking a balance between performance and efficiency, achieving state-of-the-art or comparable results in both datasets while {sharply} decreasing the number of parameters and MAdds.

\subsection{Ablation Study}

In order to investigate the effectiveness of each learning module and backbone more precisely, we conduct {an} extensive ablation study {as follows}.


\subsubsection{Different combinations of learning modules}

To demonstrate {that} all learning modules are indispensable and internally correlated, different combinations of them are compared in Table \ref{tab:module}. The models with only attractiveness distribution (AD) learning module perform poorly, with exceedingly high MAE and RMSE, along with a {very} low PC on {the} SCUT-FBP {dataset}. After introducing the rating distribution (RD) or score regression (SR) learning module, the performance significantly {improves}. {Notably, introducing} the SR learning module enables the model to perform {similarly} to the full model, which can be well explained that our task is to predict facial attractiveness in the form of {a} score. Compared with the  rating distribution, the attractiveness score is more relevant to our task. Furthermore, the performance continues to {increase} when we utilize the full model. Such growth is particularly significant on SCUT-FBP, indicating that smaller datasets benefit more from the {RD and SR} learning modules.

\begin{table*}[t]
    \centering
    \caption{Comparison of different combinations of learning modules, where AD, RD, and SR stand for the learning of attractiveness distribution, rating distribution, and score regression, respectively.}
    \label{tab:module}
    \begin{tabular}{lrrrrrr}
    \hline
    \multicolumn{1}{c}{\multirow{2}{*}{Learning module}} & \multicolumn{3}{c}{SCUT-FBP5500}                    & \multicolumn{3}{c}{SCUT-FBP}                        \\ \cmidrule(l){2-7}
    \multicolumn{1}{c}{}                         & PC$\uparrow$              & MAE$\downarrow$             & RMSE$\downarrow$            & PC$\uparrow$              & MAE$\downarrow$             & RMSE$\downarrow$            \\ \hline
    AD                                        & {0.9147{$\pm$0.0053}}          & {0.5651{$\pm$0.0001}}          & {0.6823{$\pm$0.0001}}          & {0.8168{$\pm$0.0263}}         & {0.6651{$\pm$0.0004}}          & {0.7902{$\pm$0.0005}}         \\
    AD + RD                                 & {0.9243{$\pm$0.0020}}          & {0.2094{$\pm$0.0027}}          & {0.2746{$\pm$0.0040}}          & {0.9169{$\pm$0.0047}}          & {0.2679{$\pm$0.0065}}          & {0.3301{$\pm$0.0079}}         \\
    AD + SR                                 & {0.9272{$\pm$0.0015}}          & {0.1966{$\pm$0.0022}}          & {0.2592{$\pm$0.0026}}          & {0.9271{$\pm$0.0042}}         & {0.2273{$\pm$0.0049}}         & {0.2900{$\pm$0.0073}}         \\
    AD + RD + SR                           & \textbf{0.9276{$\pm$0.0016}} & \textbf{0.1964{$\pm$0.0024}} & \textbf{0.2585{$\pm$0.0028}} & \textbf{0.9309{$\pm$0.0025}} & \textbf{0.2212{$\pm$0.0049}} & \textbf{0.2822{$\pm$0.0041}} \\ \hline
    \end{tabular}
    \end{table*}

We can draw some conclusions from the above results. First, all modules are indeed indispensable. The AD learning module is the base, and the other two refine the prediction by introducing related supervised information. To a certain extent, the SR learning module is a must for our task. Second, {the} smaller dataset requires more supervision for better performance, {and} the issue of overfitting simultaneously should be {considered}. Finally, appropriate supervision is vital for training. The choice of supervision, which avoids internal redundancy, {is also important}.

\subsubsection{Different backbones}

To explore the efficacy of different backbones, we carry out experiments under {identical} settings on {representative} backbones, which are categorized as traditional or lightweight. As shown in Table \ref{tab:backbone}, MobileNetV2 has the best overall performance on both datasets, especially on SCUT-FBP, surpassing other backbones by a large margin. {For} traditional backbones, ResNet-18 has similar results, but the parameters and MAdds {increase by} 5 times. All evaluation metrics suffer when using deeper variants of ResNet or VGG, notably in SCUT-FBP. We infer that such phenomenon mainly results from overfitting and overtraining. First, there are only 4400 and 400 training samples in {two} datasets, respectively. Training a very deep model ({e.g.,} ResNet-50) with a small or tiny dataset is prone to {overfitting}. Second, we notice that models with ResNet-50 or VGG19 have larger training {losses} than the corresponding ResNet-18 or VGG16, suggesting that the training settings might not {be} suitable for {deeper} architectures. Lastly, we adjust the training setup of {the} ResNet-50 model {so that it can be} properly trained and {achieve} improved performance; {however,} it is still inferior to the model employing MobileNetV2.


\begin{table*}[t]
    \centering
    \caption{Comparison of different backbones.}
    \label{tab:backbone}
    \resizebox{\linewidth}{!}{
    \begin{tabular}{lrrrrrrrr}
    \hline
    \multirow{2}{*}{Backbone} & \multirow{2}{*}{\#Params(M)} & \multirow{2}{*}{MAdds(G)} & \multicolumn{3}{c}{SCUT-FBP5500} & \multicolumn{3}{c}{SCUT-FBP} \\ \cmidrule(l){4-9}
                                &                              &                           & PC$\uparrow$              & MAE$\downarrow$             & RMSE$\downarrow$            & PC$\uparrow$              & MAE$\downarrow$             & RMSE$\downarrow$            \\
    \hline
    ResNet-18 & 11.20 & 1.82 & 0.9262{$\pm$0.0014} & 0.1961{$\pm$0.0020} & 0.2605{$\pm$0.0025} & 0.9238{$\pm$0.0038} & 0.2295{$\pm$0.0070} & 0.3007{$\pm$0.0082} \\
    ResNet-50 & 23.59 & 4.11 & 0.9198{$\pm$0.0016} & 0.2058{$\pm$0.0023} & 0.2728{$\pm$0.0029} & 0.9067{$\pm$0.0070} & 0.2591{$\pm$0.0090} & 0.3354{$\pm$0.0105} \\
    VGG16 & 14.74 & 15.39 & 0.9267{$\pm$0.0014} & \textbf{0.1952{$\pm$0.0019}} & 0.2593{$\pm$0.0023} & 0.9164{$\pm$0.0040} & 0.2627{$\pm$0.0121} & 0.3377{$\pm$0.0183} \\
    VGG19 & 20.06 & 19.55 & 0.9251{$\pm$0.0014} & 0.1978{$\pm$0.0021} & 0.2621{$\pm$0.0023} & 0.9135{$\pm$0.0148} & 0.2719{$\pm$0.0258} & 0.3515{$\pm$0.0459} \\
    \hline
    MobileNetV3\_large & 4.25 & 0.22 & 0.9210{$\pm$0.0018} & 0.2025{$\pm$0.0022} & 0.2688{$\pm$0.0028} & 0.9122{$\pm$0.0057} & 0.2776{$\pm$0.0106} & 0.3472{$\pm$0.0128} \\
    MobileNetV3\_small & 1.56 & 0.06 & 0.9103{$\pm$0.0019} & 0.2142{$\pm$0.0023} & 0.2850{$\pm$0.0028} & 0.9156{$\pm$0.0059} & 0.2631{$\pm$0.0099} & 0.3427{$\pm$0.0148} \\
    MobileNetV2 & 2.28 & 0.31 & \textbf{0.9276{$\pm$0.0016}} & 0.1964{$\pm$0.0024} & \textbf{0.2585{$\pm$0.0028}} & \textbf{0.9309{$\pm$0.0025}} & \textbf{0.2212{$\pm$0.0049}} & \textbf{0.2822{$\pm$0.0041}} \\
    \hline
    \end{tabular}}
\end{table*}

{For} lightweight backbones, we conduct experiments on MobileNetV3~\cite{howard2019searching}. The formerly proposed MobileNetV2 still enjoys clear superiority. We notice that the performance declines with larger MobileNetV3 on SCUT-FBP, which is consistent with the traditional backbones. Thus, we can conclude that the choice of backbone in terms of scale and structure is {significant in} performance and requires careful consideration.


\subsubsection{Different label distribution learning schemes}

To demonstrate the superiority of our approach among LDL methods, we carry out the following comparative experiments. The approach in \cite{fan2017label} is reimplemented on SCUT-FBP5500, since the dataset had not been released at the time {of publication of} the paper. Then, the method in \cite{gao2018age}, which was originally designed for age estimation, was adapted to FAP {since it is} a similar regression task {to that used} in \cite{gao2020learning}. For {a} fair comparison, the methods mentioned above are reimplemented using MobileNetV2 under identical training settings. Furthermore, we compare the {performances} of {the} Gaussian and Laplace {distributions} by replacing the probability distribution employed in the attractiveness distribution. As Table \ref{tab:LDL} {shows}, our delicately designed approach achieves the best performance on both datasets, notably on SCUT-FBP, {greatly} outperforming {the} other methods. When comparing Laplace and Gaussian {distributions}, our approach with {the} Laplace distribution performs slightly better on SCUT-FBP5500 {but significantly better} on SCUT-FBP, proving the effectiveness and simplicity of Laplace distribution.

\begin{table*}[t]
    \centering
    \caption{Comparison of different LDL schemes.}
    \label{tab:LDL}
    \begin{tabular}{@{}lrrrrrr@{}}
    \hline
    \multirow{2}{*}{LDL scheme}    & \multicolumn{3}{c}{SCUT-FBP5500}                    & \multicolumn{3}{c}{SCUT-FBP}                        \\ \cmidrule(l){2-7}
                                                                & PC$\uparrow$              & MAE$\downarrow$             & RMSE$\downarrow$            & PC$\uparrow$              & MAE$\downarrow$             & RMSE$\downarrow$            \\ \hline
    \cite{fan2017label} & 0.9259{$\pm$0.0018} & 0.1965{$\pm$0.0026} & 0.2606{$\pm$0.0032} & 0.9211{$\pm$0.0058} & 0.2283{$\pm$0.0095} & 0.2954{$\pm$0.0100} \\
    \cite{gao2020learning} & 0.9253{$\pm$0.0016} & 0.2074{$\pm$0.0019} & 0.2722{$\pm$0.0024} & 0.9184{$\pm$0.0040} & 0.2692{$\pm$0.0029} & 0.3439{$\pm$0.0042}\\
    Ours with Gaussian distribution & 0.9275{$\pm$0.0014} & 0.1966{$\pm$0.0021} & 0.2587{$\pm$0.0024} & 0.9263{$\pm$0.0040} & 0.2271{$\pm$0.0063} & 0.2910{$\pm$0.0075} \\
    Ours with Laplace distribution  & \textbf{0.9276{$\pm$0.0016}} & \textbf{0.1964{$\pm$0.0024}} & \textbf{0.2585{$\pm$0.0028}} & \textbf{0.9309{$\pm$0.0025}} & \textbf{0.2212{$\pm$0.0049}} & \textbf{0.2822{$\pm$0.0041}} \\ \hline
    \end{tabular}
\end{table*}

\subsection{Visualization}

To better understand how our model {perceives abstract} facial attractiveness, we visualize a feature map {that} can intuitively present the prediction patterns of different degrees of attractiveness, and indicate the reasons {for} good or poor predictions.

\begin{figure}[t]
    \centering
    \includegraphics[width = .9\linewidth]{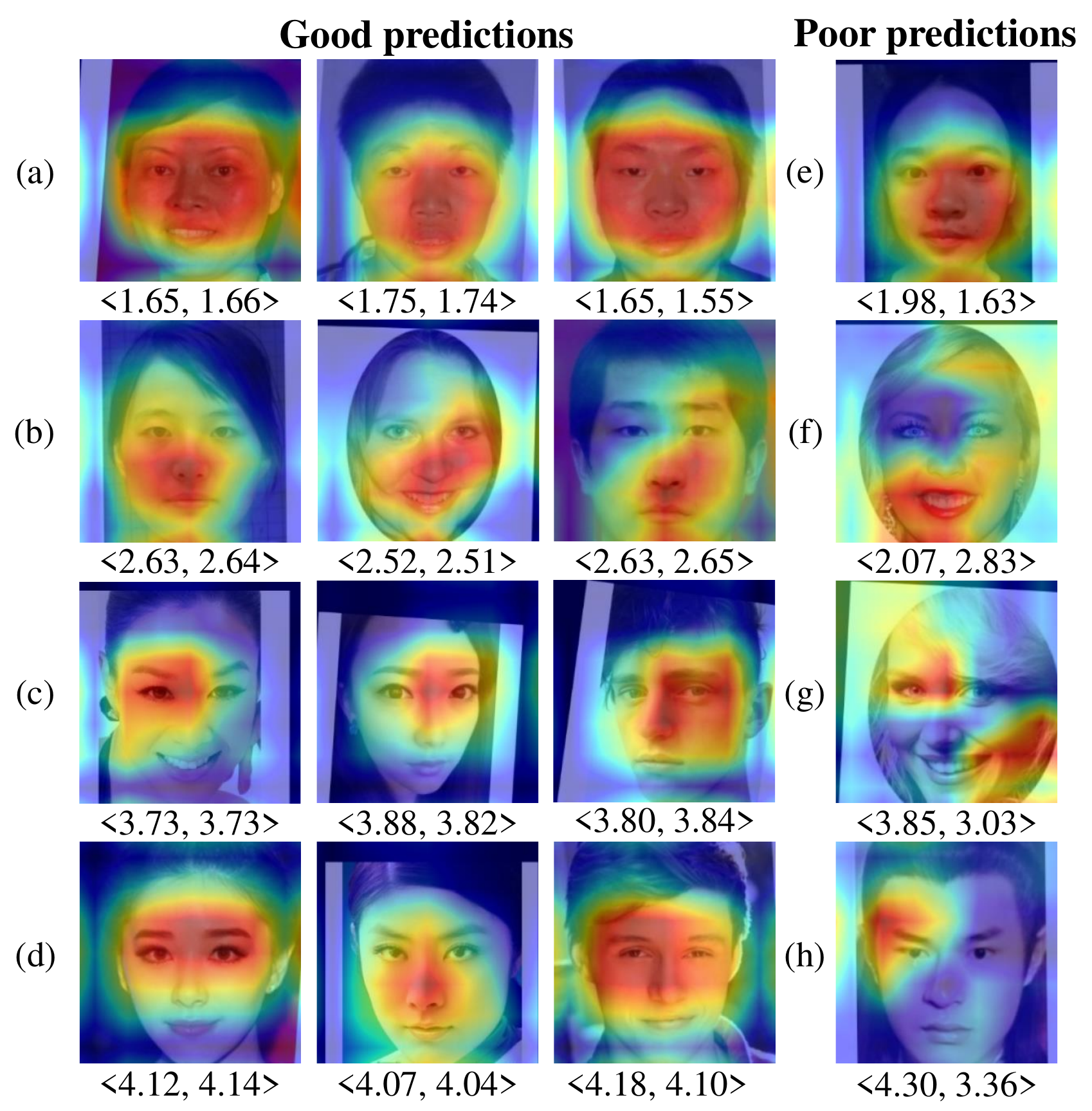}
    \caption{The heatmap visualization, where warmer colors (e.g., red) indicate higher {intensities} and cooler colors (e.g., blue) indicate lower {intensities}. Each row {corresponds to} a distinct degree of attractiveness. The left three columns ((a)-(d)) and the rightmost column ((e)-(h)) are examples of good and poor predictions, respectively. The pairs below the images {represent the} $<$ground-truth score, predicted score$>$.}
    \label{fig:heatmap}
\end{figure}

Here the visualization follows the class activation mapping (CAM) method \cite{zhou2016learning}. The global average pooling (GAP) is applied to the last feature map of the model to calculate the mean of each channel. The results are then mapped to the attractiveness score via the fully-connected layer, and its gradients with respect to the last feature map {are} calculated. The ReLU6 activation layer in the last convolution block of MobileNetV2 produces 7$\times$7 feature maps with 1280 channels. These feature maps are first channel-wise averaged and resized to 224$\times$224. Finally, those gradients are visually presented on the input images, as shown in Fig.~\ref{fig:heatmap}.

We observe that our approach shows different perception patterns for faces with different levels of attractiveness. When the ground-truth score is low, the approach evaluates the face by looking at the larger facial areas. It is consistent with findings in cognitive neuroscience and clinical psychology, that people tend to devote more attention to full observation of the face before giving {a} low rating \cite{richards2014exploring,zhu2022preference}. With {an} increasing degree of attractiveness, our approach gradually focuses on certain facial regions (e.g., mouth, eyes, nose) and produces semantic predictions. It is consistent with findings that people are easily attracted to delicate features (e.g., sensuous lips, large eyes, high nose), and thus give high {scores to faces} with such positive characteristics \cite{kou2016attentional,zhu2022preference}.

From good predictions, human-like behavior can be discovered to demonstrate the effectiveness of our approach. In Fig.~\ref{fig:heatmap}(a), the {high-intensity} areas almost cover the whole face, indicating that the model fails to locate attractive facial regions, resulting in a relatively low attractiveness score. In Fig.~\ref{fig:heatmap}(b)(c), the {high-intensity} areas gradually narrow down to smaller areas, presenting ambiguously semantic predictions. Specifically, when the face is attractive, the model is able to capture {abstract} facial attractiveness, concentrating on facial regions with evident semantics, such as the eyes and nose in Fig.~\ref{fig:heatmap}(d). From poor predictions, however, we notice that our approach mainly focuses on {irrelevant} regions, such as hair, regions outside face, and background, leading to failure in capturing attractiveness, as shown in Fig.~\ref{fig:heatmap}(e)-(h).

{In summary}, our approach is capable of sensing facial beauty and capturing attractive facial regions to accomplish accurate and efficient FAP. However, it still has some limitations, mainly due to focusing on {irrelevant} areas of the images.

\subsection{Discussion}
Here we highlight five observations from the experiments. Compared with the state-of-the-art works, our method shows advantages in both performance and efficiency. Compared with the traditional single label, our dual label distribution is conducive to improving the FAP performance, especially on smaller-scale datasets. Our well-designed learning modules of attractiveness distribution, rating distribution and score regression play different roles, but they are all indispensable and intrinsically related. Compared with large-scale backbones, {lightweight models can} reduce the risk of overfitting and perform better on small datasets. By investigating the probability distribution in label distribution construction, the advantages of modeling discrete labels with continuous distribution functions over discrete distributions and using {the Laplace} distribution in our approach over {the} Gaussian can be observed.

However, some limitations {to} our work {remain}. First, some failure cases exist, where our approach identifies {facial} features less pertinent to perceived attractiveness, leading to poor predictions. Second, our work is based on static images {and} fails to {consider} temporal cues. In fact, psychological \cite{rubenstein2005variation} and neuroscience \cite{o2002recognizing} studies have proven that temporal cues play a vital role in perceiving facial attractiveness. Nevertheless, little attention has {been given} to dynamic facial {content}, and the temporal dynamics of facial attractiveness {remain} largely unexplored. Kalayci \textit{et al.} \cite{kalayci2014automatic} utilized dynamic and static features extracted from video clips for facial attractiveness analysis. Recently, Weng \textit{et al.} \cite{weng2021two} conducted dynamic facial attractiveness prediction utilizing videos from TikTok. Third, only frontal facial images are employed in our work, which lack variations in visual angles so that fail to reflect the true {facial} structure. Researchers have shown that facial attractiveness is jointly determined by {the} frontal view, profile view, and their combination \cite{liao2012enhancing}. Therefore, adopting multi-view or three-dimensional facial data for facial attractiveness analysis and prediction, which is expected to produce more comprehensive and reliable results and thus better reveal the secrets of facial attractiveness, {is important}. However, such direction has received little attention, and it was {not until recently} that some researchers started to investigate \cite{liu2017landmark,xiao2021beauty3dfacenet,liu2022computation}.

\section{Conclusion}

In this paper, we integrate the lightweight design and LDL paradigm to develop a novel facial attractiveness prediction model {that consists of} (1) {a dual-label} distribution to take full advantage of the dataset, {and} (2) a joint learning framework to optimize the dual label distribution and attractiveness score simultaneously. The proposed approach achieves appealing results with greatly decreased parameters and computation. The visualization demonstrates that our approach is interpretable, employing different patterns to capture facial attractiveness so as to generate semantic predictions.

This work {has} several interesting directions for future {exploration}. The proposed {dual-label} distribution is expected to generalize to other similar tasks, such as age prediction and facial expression recognition. For datasets without sufficient information, pseudo distribution can be generated to employ our dual {LDL} paradigm. In addition, customizing {facial} attractiveness prediction models is more applicable to person-specific {situations}, like online dating recommendation. Such personalized prediction leverages previous ratings from a target individual or others with similar preferences to develop a customized model, {which is} potentially more challenging. The body of related studies is small {and} the reported accuracy is inferior to universal models.


\bibliographystyle{IEEEtran}
\bibliography{IEEEabrv,ref_20240201}

\end{document}